\documentclass[journal,transaction]{IEEEtran}
\usepackage{graphicx}
\usepackage{subfigure}
\usepackage{amsmath}
\usepackage{color}
\usepackage{cite}
\usepackage{amsmath,amssymb,amsfonts}
\usepackage{algorithm}
\usepackage{algorithmic}
\usepackage{textcomp}
\usepackage{booktabs}
\usepackage{epstopdf}
\usepackage{array}
\usepackage{pgffor}
\usepackage{threeparttable}
\usepackage{multicol}
\usepackage{mathtools}
\usepackage[marginal]{footmisc}
\usepackage{arydshln}

\usepackage[colorlinks,linkcolor=green]{hyperref}

\newcommand{\fig}[1]{\mbox{Figure~\ref{#1}}}
\newcommand{\eq}[1]{\mbox{Eq.~(\ref{#1})}}

\newcommand{\tab}[1]{\mbox{Table~\ref{#1}}}

\begin{document}

\title{Study on a Fast Solver for Combined Field Integral Equations of 3D Conducting Bodies Based on Graph Neural Networks}

\author{\IEEEauthorblockN{Tao Shan,~\IEEEmembership{Member,~IEEE}, Xin Zhang, and Di Wu}
	\thanks{This work was supported in part by the National Natural Science Foundation of China under Grant 62401035; in part by School of Electronic Information Engineering, Beihang University; in part by Ministry of Industry and Information Technology Key Laboratory of Electromagnetic Environment Effect of Intelligent Systems, Beihang University.(Corresponding author: Tao Shan.)}
	\thanks{Tao Shan, Xin Zhang, and Di Wu are with School of Electronics and Information Engineering, Beihang University, Beijing 100191, China.\\
	(e-mail:taoshan@buaa.edu.cn).}}
\maketitle
\begin{abstract}
	In this paper, we present a graph neural networks (GNNs)-based fast solver (GraphSolver) for solving combined field integral equations (CFIEs) of 3D conducting bodies.
	Rao-Wilton-Glisson (RWG) basis functions are employed to discretely and accurately represent  the geometry of 3D conducting bodies.
	A concise and informative graph representation is then constructed by treating each RWG function as a node in the graph, enabling the flow of current between nodes. 
	With the transformed graphs, GraphSolver is developed to directly predict real and imaginary parts of the x, y and z components of the surface current densities at each node (RWG function).
	Numerical results demonstrate the efficacy of GraphSolver in solving CFIEs for 3D conducting bodies with varying levels of geometric complexity, including basic 3D targets, missile-shaped targets, and airplane-shaped targets.
\end{abstract}
\section{Introduction}
Electromagnetic (EM) scattering serves as a common concept in electrical engineering\cite{chew2001fast, gibson2021method, jin2011theory, chew2022integral}, underpinning diverse applications such as microwave imaging\cite{pastorino2018microwave}, geophysical exploration\cite{abubakar20082}, remote sensing\cite{jin1993electromagnetic}, and electromagnetic compatibility\cite{paul2022introduction}.
Among various scattering targets, 3D conducting bodies represent a critical category due to their extensive use in practical applications, including aircraft and ships\cite{jin2011theory, chew2022integral}.
Many research efforts have been dedicated to modeling 3D conducting bodies using techniques such as the finite difference method\cite{taflove2005computational}, finite element method\cite{jin2015finite}, method of moments (MoM)\cite{gibson2021method}, and discontinuous Galerkin time-domain methods\cite{chen2012discontinuous}.
These bodies often exhibit intricate geometries, encompassing multiple scales and detailed structures.
Consequently, the associated matrix equations typically involve a large number of unknowns, rendering accurate electromagnetic modeling both computationally intensive and time-consuming.
To address these challenges, fast algorithms have been developed to mitigate computational complexity and accelerate calculations, including adaptive cross-approximation\cite{zhao2005adaptive}, conjugate gradient–fast Fourier transform\cite{sarkar1986application}, fast multipole algorithm\cite{rokhlin1985rapid}, etc.
Despite these advancements, real-time and reliable modeling remains a persistent issue in the electromagnetic analysis of 3D conducting bodies.
\par 
Recent advancements in deep learning (DL) have significantly enhanced computational efficiency in electromagnetics\cite{erricolo2019machine,massa2019dnns,salucci2022artificial,chen2020review}, finding applications in areas such as electromagnetic modeling\cite{yao2018machine,shan2020study,shan2021application,9206148,zhu2024surrogate,cao2024raypronet,ooi2022multi,ma2020learning,key2021predicting,10508749,qi2024hybrid,guo2021electromagnetic,hu2021theory,shan2023physics,shan2023solving}, microwave imaging\cite{li2018deepnis,song2021electromagnetic,khoshdel2023multi,10041848,dai20233dinvnet,shan2022neural,zhang2022unrolled}, and inverse design\cite{shan2021phase,shan2020coding,smith2023real,feng2022artificial}. 
Data-driven learning presents a straightforward and effective approach to incorporating DL in electromagnetics\cite{yao2018machine,shan2020study,shan2021application,9206148,zhu2024surrogate,cao2024raypronet,ooi2022multi,ma2020learning,key2021predicting,10508749}.
The core concept is to directly establish nonlinear mappings between various physical quantities by training deep neural networks (DNNs) to extract physical laws from vast datasets.
In this context, DNNs serve as surrogate models, offering reduced computational complexity compared to traditional, computationally expensive algorithms.
Although the offline training of DNNs can be time-consuming, online computations are substantially accelerated through GPU parallelization. 
The training process typically requires large volumes of data, with the quality of this data directly impacting the performance of the DNNs. 
However, the lack of interpretability remains a key limitation, especially in applications where reliability is paramount.
\par 
To enhance both interpretability and robustness, physics-inspired learning has emerged, integrating the mathematical relationships between DNNs, electromagnetic physics, and numerical algorithms. 
Maxwell’s equations, along with various boundary conditions, can act as effective physical constraints to guide DNN training\cite{qi2024hybrid}.
The mathematical similarities between the finite-difference time-domain method and recurrent neural networks (RNNs) or convolutional neural networks (CNNs) have been identified and leveraged\cite{guo2021electromagnetic,hu2021theory}.
Additionally, iterative solvers can train DNNs as learned parametric update functions, improving convergence while preserving the essential components of numerical computations\cite{shan2023physics,shan2023solving}.
In the context of microwave imaging, iterative optimization algorithms can be unrolled into DNNs by treating each iteration as a distinct layer\cite{shan2022neural,zhang2022unrolled}. Furthermore, DNNs can approximate mathematical operators, learning mappings between infinite-dimensional spaces, and thus eliminating dependencies on mesh densities and domain shapes\cite{li2020fourier}.
\par
Despite the successful application of DL in electromagnetic, much of the published research has focused primarily on solving 2D and 3D EM modeling problems using uniform meshes.
However, nonuniform meshes are often the preferred choice for accurately modeling 3D electromagnetic problems. Most DL techniques, which are designed for structured data, are not directly applicable to processing nonuniform meshes.
To the best of the author's knowledge, a few DL techniques have been applied to 3D EM modeling.
Point cloud, which is an effective method for describing complex 3D geometries in DL, has been used to represent the geometric information of 3D PEC targets by employing individual points in 3D space instead of nonuniform meshes \cite{10508749}.
PointNet has been trained to extract feature parameters for calculating the scattered far fields, taking the point cloud of the PEC target as input \cite{10508749}.
Despite its impressive performance, point cloud describes the shape of an object without considering the relationships between these points, rendering it unsuitable for surface meshes.
Graph neural networks (GNNs), on the other hand, are well-suited for processing graph-structured, unstructured data.
Inspired by this, PhiGRL employs GNNs for 3D EM modeling by transforming nonuniformly discretized 3D PEC targets into graphs\cite{shan2023physics}.
In this approach, PhiGRL trains GNNs to iteratively adjust the candidate solution until convergence.
However, because PhiGRL incorporates EM physics, it places higher demands on computational resources, which makes it less efficient for training compared to data-driven models.
\par 
In this paper, we propose a graph neural networks-based solver to solve CFIEs of 3D conducting bodies.
A concise and informative graph representation of 3D conducting bodies is introduced by transforming the applied RWG basis functions into graphs.
This representation treats each triangular patch as a node in the graph, allowing the RWG functions to enable the flow of current between nodes \cite{chew2022integral}.
GraphSolver is then developed to directly predict real and imaginary parts of the x, y and z components of surface current densities by processing graph data derived from 3D conducting bodies. 
The architecture of GraphSolver sequentially employs an upsampling fully connected network (FCN), a graph convolutional network (GCN), and six downsampling FCNs.
The efficacy of GraphSolver is validated by solving CFIEs for 3D conducting bodies with varying levels of geometric complexity, including basic 3D targets, missile-shaped targets, and airplane-shaped targets.
The implementation code and trained model parameter files for basic 3D targets, as well as missile-shaped and airplane-shaped targets, are publicly available on the website \href{https://github.com/IEMCS-Lab/GNNsolver-CFIE/}{https://github.com/IEMCS-Lab/GNNsolver-CFIE/} for access and use.
\par 
This paper is organized as follows. Section \uppercase\expandafter{\romannumeral2} reviews the combined field integral equation.
Section \uppercase\expandafter{\romannumeral3} formulates the proposed fast solver based on graph neural networks.
In Section \uppercase\expandafter{\romannumeral4}, the proposed fast solver is applied to separately solve CFIEs of 3D conducting bodies exhibiting varying degrees of geometric complexity. 
Observations and discussions are summarized in Section \uppercase\expandafter{\romannumeral5}.
\section{Combined Field Integral Equation}
The combined-field integral equation characterizes the surface currents on a conducting body with a closed surface $S_0$ under illumination by a specified incident wave\cite{chew2001fast,jin2011theory,gibson2021method}:
\begin{equation}
	\begin{split}
		 \alpha & \cdot Z_0 \cdot   [\frac{1}{2}{\bf{J} _{s}(\bf{r})} + \hat n \times \overline{\overline{\cal K}} ({{\bf{J}}_{s}(\bf{r})}) - \hat n \times  {\bf{H}}^{inc}({\bf{r}})] \\
		& -(1-\alpha) \cdot \hat n \times \hat n \times[ \overline{\overline{\cal L}}( Z_0{\bf{J}}_{s}({\bf{r}}) ) - {\bf{E}}^{inc}({\bf{r}})] =0
		\label{cfie}
	\end{split}
\end{equation}
where $\bf{r} \in S_{0}$ and $ \hat n$ are the position vector and normal vector of $S_{0}$, ${\bf{E}}^{inc}$ and $\mathbf{H}^{inc}$ denote the incident electric and magnetic fields, $\mathbf{J}_{s}(\mathbf{r})$ is the surface current and $Z_0$ is the wave impedance.
$\overline{\overline{\cal L}}( \cdot )$ and 
$\overline{\overline{\cal K}}( \cdot )$ are the integral operators and their definitions can refer to \cite{chew2001fast,jin2011theory,gibson2021method}.
\par
\begin{figure}
	\centering
	\subfigure[Vector plot of RWG basis function]{\includegraphics[width=0.7\linewidth]{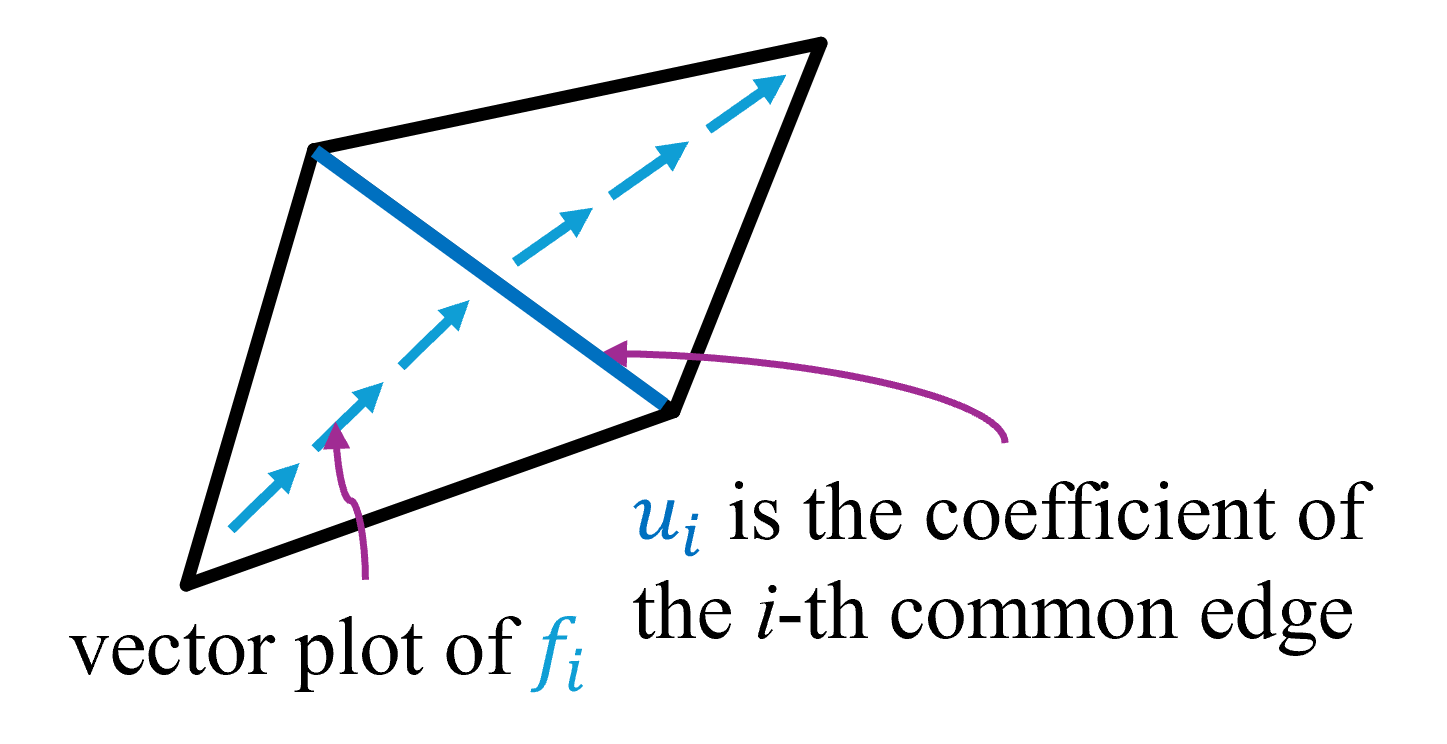}
		\label{RWG1}}
	\subfigure[Surface current density of a single triangular element]{\includegraphics[width=0.7\linewidth]{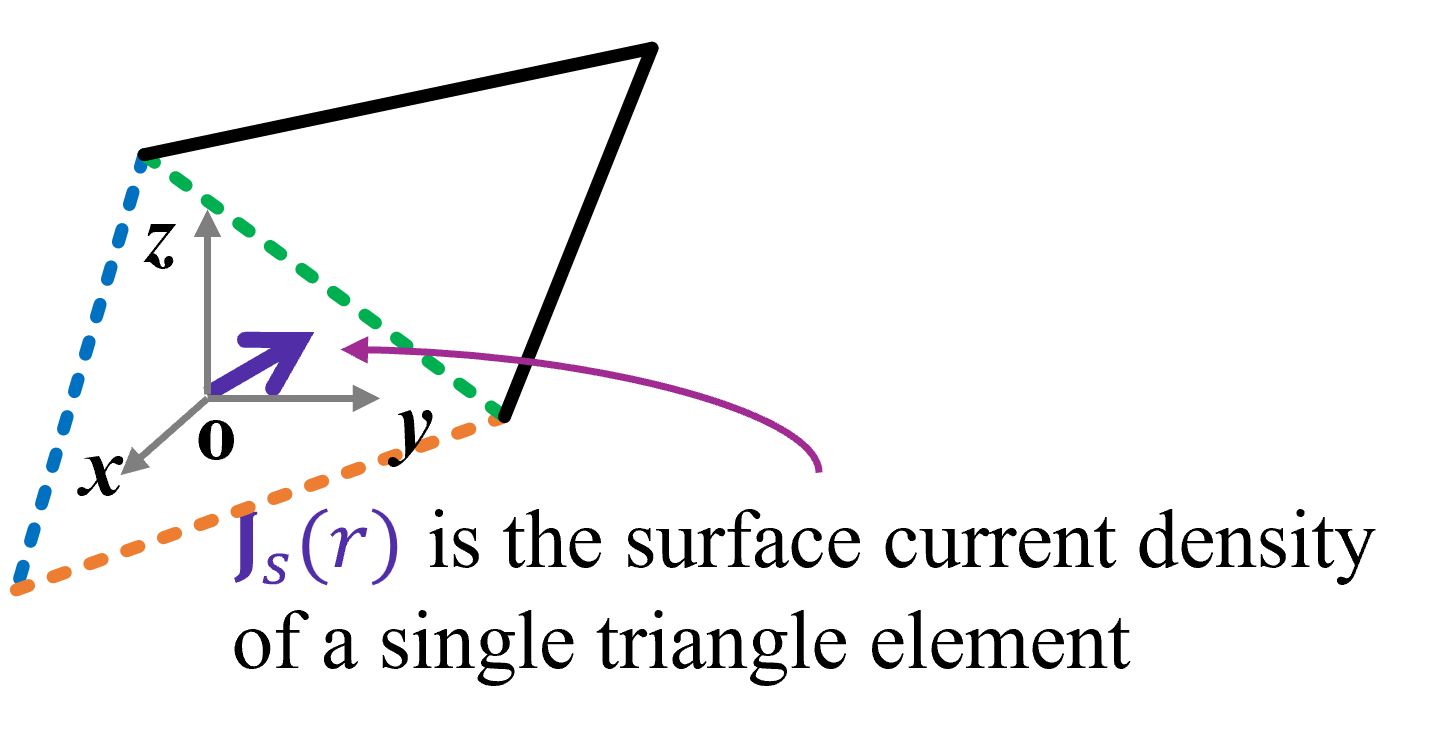}
		\label{RWG2}}
	\caption{Relationship between RWG basis function and surface current density.}
	\label{RWG}
\end{figure}

 
MoM is an efficient approach for solving \eq{cfie} by representing the surface current density, $\mathbf{J}_{s}(\mathbf{r})$, with Rao-Wilton-Glisson (RWG) basis functions\cite{chew2001fast,jin2011theory,gibson2021method}, as shown in \fig{RWG}:
\begin{equation}
	\mathbf{J}_{s}(\mathbf{r}) = \sum_{i=1}^{N} u_i f_i(\mathbf{r})
	\label{RWGeq}
\end{equation}
where $f_i(\mathbf{r})$ denotes the $i$-th RWG basis functions defined over two triangular elements sharing a common edge, and $u_i$ is the corresponding coefficient of $f_i(\mathbf{r})$. Consequently, \eq{cfie} can be converted into a matrix equation:
\begin{equation}
	\mathbb{Z} \cdot \mathbf{u} = \mathbf{b}
	\label{matrixeq}
\end{equation}
where $\mathbb{Z}$ and $\mathbf{b}$ are the impedance matrix and excitation vector, $\mathbf{u}$ is the vector of coefficients.
\begin{figure*}
	\centering
	\includegraphics[width=1\linewidth]{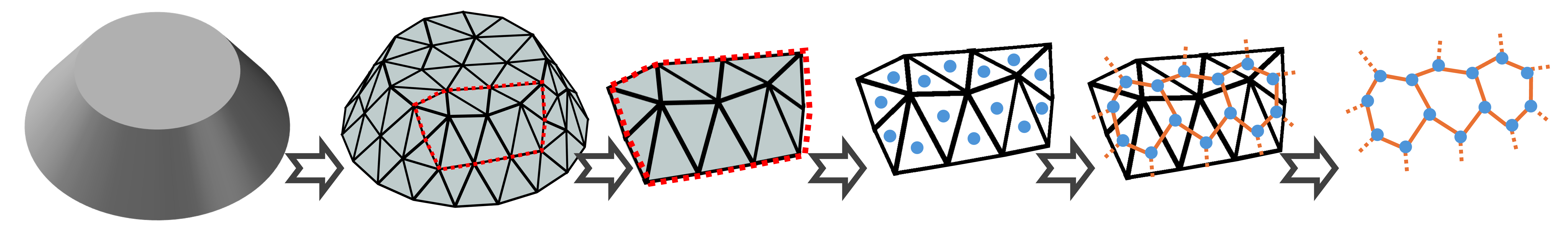}
	\caption{Graph representation of RWG basis functions in a 3D conducting body. The 3D conducting body is first discretized into  triangular meshes, with each triangular element represented as a node in the graph. Two nodes are connected to form an edge in the graph if their corresponding triangular elements share a common side.}
	\label{GraphRep}
\end{figure*}
\begin{figure*}
	\centering
	\includegraphics[width=1\linewidth]{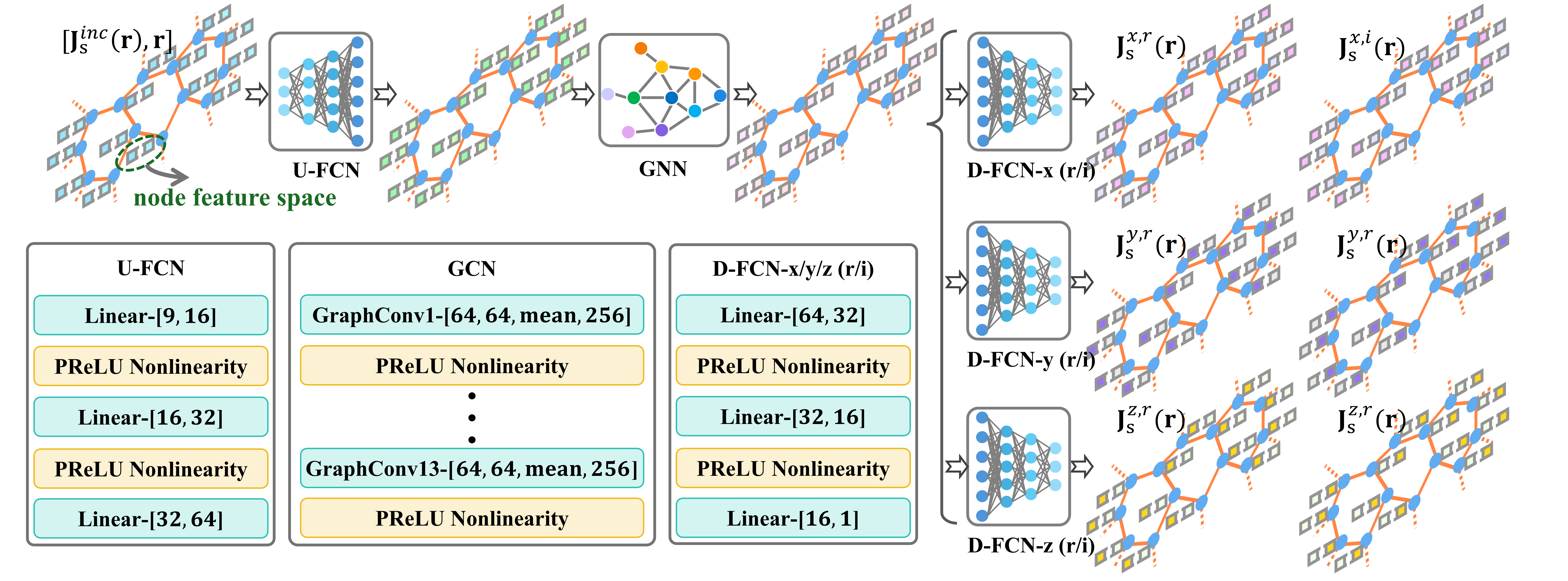}
	\caption{GraphSolver for solving CFIEs of 3D conducting bodies. It consists of one upsampling FCN, one GCN, and six downsampling FCNs. Their detailed architectures are illustrated. U-FCN is the upsampling FCN. D-FCN-x/y/z(r/i) denotes the downsampling FCN for the real (r) and imaginary (i) parts of the x, y and z components of the surface current density. Linear-$[a,b]$ denotes a linear layer with the input and output channels $a$ and $b$ respectively. GraphConv$k-[c,d,mean,w]$ denotes the $k$-th graph convolutional layer of which the input channel, output channel, aggregation function and kernel width are $c$, $d$, mean function and $w$.}
	\label{GNN}
\end{figure*}
\section{Fast Solver based on Graph Neural Network}
In EM modeling of 3D conducting bodies, triangular mesh is an effective choice for accurately capturing complex geometries with controllable precision.
However, this approach introduces unstructured data, as the number and distribution of meshes vary across different conducting bodies. 
This variability limits the applicability of most deep learning techniques, which are typically designed for structured data. 
Graph neural networks, a class of artificial neural networks tailored for graph-structured data, offer a solution for processing this unstructured data\cite{li2020fourier}. 
Inspired by this, we propose a concise and informative graph representation for 3D conducting bodies in EM modeling. 
Then, a fast GNN-based solver is developed to address the CFIE of 3D conducting bodies using these graph representations.
\subsection{Graph Representation of 3D conducting bodies}
In MoM, a 3D conducting body is initially divided into triangular patches.
The surface current density $\mathbf{J}_{s}(\mathbf{r})$ is then expanded with a set of RWG basis functions, as formulated in \eq{RWGeq}.
\fig{RWG1} depicts the vector plot of the RWG basis function.
The $i$-th RWG function $f_i$ is defined over two triangular elements sharing the $i$-th common edge, with its coefficient $u_i$ associated with this edge.
Consequently, the number of unknowns corresponds to the number of edges present in all triangular elements.
After solving \eq{matrixeq} for $\mathbf{u}$, the surface current density $\mathbf{J}_{s}(\mathbf{r})$ can be determined based on \eq{RWGeq}, as shown in \fig{RWG2}.
\par
Unlike MoM, this paper aims to employ GNNs to directly predict the surface current density $\mathbf{J}_{s}(\mathbf{r})$ of each triangular element instead of determining the coefficient $\mathbf{u}$ of RWG functions, as shown in \fig{RWG2}.
Before applying GNNs, a graph representation is constructed from the RWG basis functions that define the 3D conducting body.
A graph is defined as a pair of two sets $G=(V,E)$ where
$V=\{v_i|i=1, \cdots, M\}$ is a finite set of $M$ nodes and $E=\{e_{ij}|e_{ij}=(v_i, v_j)\in V^2, v_i\neq v_j\}$  represents the edges connecting these nodes.
Each node $v_i$ in $V$ can have its own feature vector $F(v_i)$.
\fig{GraphRep} illustrates the process of transforming RWG basis functions into a graph.
First, a 3D conducting body is discretized into triangular elements, then each triangular element is treated as a node in the graph. Two nodes are connected by an edge if their corresponding triangular elements share the same side.
In this way, a graph representation $G=(V,E)$ of a 3D conducting body is established.
It should be noted that the number of nodes in $V$ equals the number of triangular elements.
The proposed graph representation not only preserves the geometric information of the 3D conducting body but also adapts to variations in the number and distribution of unknowns across different 3D conducting structures.
This representation is a viable option, as it is straightforward, concise and informative.
\begin{table}
	\caption{Parameter of the kernel function $\mathcal{K}_{\mathbf{W}}$}
	\label{table01}
	\centering
	\begin{tabular}{|c|c|}
		\hline
		Module 1&Linear-[3, 256], PReLU nonlinearity \\ \hline
		Module 2&Linear-[256, 256], PReLU nonlinearity\\ \hline
		Module 3&Linear-[256, 256], PReLU nonlinearity \\ \hline
		Output layer&Linear-[256, 4096] \\ \hline
	\end{tabular}
	\begin{tablenotes}
		\item Linear-$[a,b]$ denotes a linear layer with the input and output channels $a$ and $b$ respectively.
	\end{tablenotes}
\end{table}
\subsection{Network Architecture}
In this section, GraphSolver is developed as a fast solver for CFIEs of 3D conducting bodies by directly predicting surface current densities.
\fig{GNN} illustrates the architecture of GraphSolver, which sequentially employs an upsampling FCN, a GCN, and six downsampling FCNs.
Six downsampling FCNs are separately trained to predict the real (r) and imaginary (i) parts of the x, y and z components of the surface current density. 
Their detailed architectures are also present in \fig{GNN}.
\par 
The graph representation of a 3D conducting body can be denoted as $G=(V,E,F)$, where $V$ denotes the triangular elements, $E$ represents their connections, and $F$ includes the feature vectors associated with all triangular elements.
Both input and output of the proposed GNN model are the graph representations.
These graph representations share an identical structure defined by $V$ and $E$, but they differ in feature vectors $F$.
This setup ensures that the geometric information of the 3D conducting bodies is consistently incorporated in the computations.
This approach aligns with the goal of EM modeling, which is to compute the physical quantities for each triangular element.
In this respect, the proposed GNN model demonstrates a distinct advantage.
\par
The input graph can be represented as $G_{in}=(V,E,F_{in})$ with $M$ nodes. The feature vector of $i$-th node, $F_{in}(v_i) = [\mathbf{J}_{s}^{inc}(\mathbf{r}), \mathbf{r}] \in \mathbb{R}^9$, comprises the surface current density induced by the incident electric and magnetic fields, as well as the position vector of the $i$-th triangular element:
\begin{equation}
	\mathbf{J}_{s}^{inc}(\mathbf{r}) = \hat n \times  Z_0 {\bf{H}}^{inc}({\bf{r}}) - \hat n \times \hat n \times {\bf{E}}^{inc}({\bf{r}})
\end{equation}
To enhance the feature vector dimensions in $G_{in}=(V,E,F_{in})$, an upsampling FCN is first applied:
\begin{equation}
	G_{up} = (V,E,F_{up}) = \Phi_{up}(G_{in}, \Theta_{up})
\end{equation}
where $F_{in}\in \mathbb{R}^{M\times 9}$, $F_{up}\in \mathbb{R}^{M\times 64}$, $\Phi_{up}$ and $\Theta_{up}$ denote the upsampling FCN and its parameter set. 
Next, a GCN is applied to $G_{up}$:
\begin{equation}
	G_{gcn} = (V,E,F_{gcn}) = \Phi_{gcn}(G_{up}, \Theta_{gc})
\end{equation}
where $F_{gcn}\in \mathbb{R}^{M\times 64}$, $\Phi_{gcn}$ and $\Theta_{gcn}$ denote the GCN and its parameter set. The graph convolution operation, as defined in\cite{gilmer2017neural}, is given by:
\begin{equation}
	F(v_i^{p+1}) = \mathbf{W} F(v_i^{p}) + \frac{1}{N_{\mathcal{N}(i)}}
	\sum_{j \in \mathcal{N}(i)} F(v_j^{p}) \cdot
	\mathcal{K}_{\mathbf{W}}(\mathbf{r}_{\mathbf{e}_{i,j}})
\end{equation}
where $F(v_i^{p})$ and $F(v_i^{p+1})$ denote the feature vectors of the $i$-th node at the $p$-th and $p+1$-th layer, $\mathcal{K}_{\mathbf{W}}$ is a trainable kernel function, $\mathcal{N}(i)$ is the set of adjacent nodes of the $i$-th node, $N_{\mathcal{N}(i)}$ is the number of adjacent nodes, $\mathbf{W}$ is the trainable weight matrix, and $\mathbf{r}_{\mathbf{e}_{i,j}}$ is the position vector of the edge $\mathbf{e}_{i,j}$.
The kernel function $\mathcal{K}_{\mathbf{W}}$ in this paper is implemented as an FCN, with its detailed structure provided in \tab{table01}.
Finally, six downsampling FCNs are trained independently to predict the real (r) and imaginary (i) parts of the x, y and z components of the surface current density based on $G_{gcn}$:
\begin{equation}
	\begin{split}
	G_{x/y/z,r/i} &= (V,E,F_{x/y/z,r/i}) \\
	&= \Phi_{x/y/z,r/i}(G_{gcn}, \Theta_{x/y/z,r/i})
	\end{split}
\end{equation}
where $F_{x/y/z,r/i} \in \mathbb{R}^{M\times1}$, $\Phi_{x/y/z,r/i}$ and $\Theta_{x/y/z,r/i}$ are the downsampling FCNs and their respective parameter sets that are applied to predict the real (r) and imaginary (i) parts of the x, y and z components of $\mathbf{J}_{s}(\mathbf{r})$.
\begin{figure}
	\centering
	\includegraphics[width=1\linewidth]{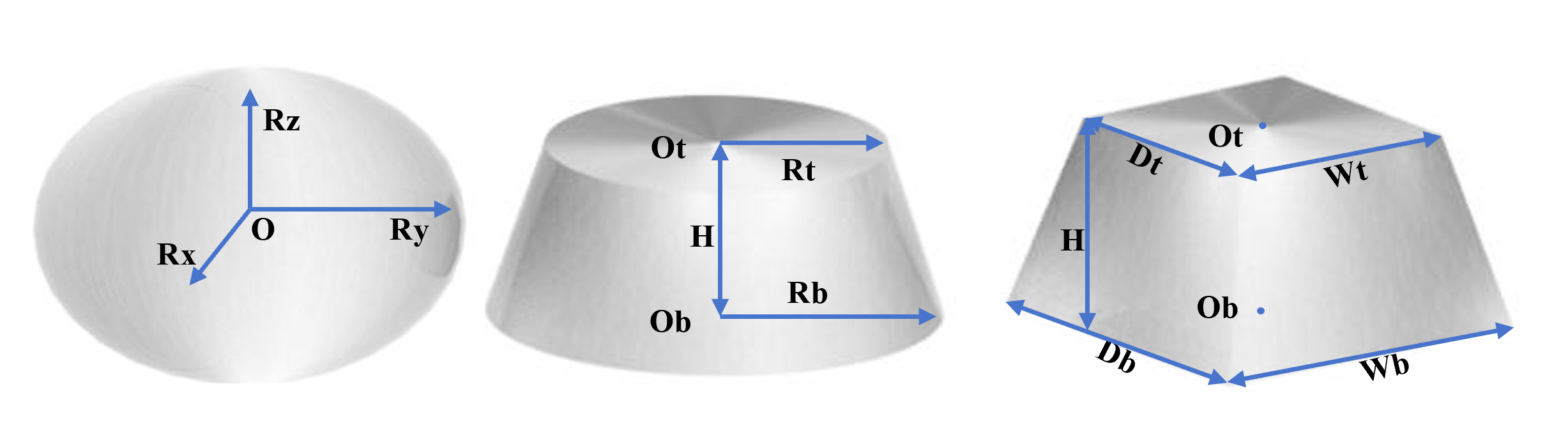}
	\caption{Schematic of basic 3D targets: spheroid, conical frustum, and hexahedron (from left to right). O, Ot and Ob denotes the body center, top center and base center.}
	\label{basictargets}
\end{figure}
\begin{figure}
	\centering
	\includegraphics[width=0.85\linewidth]{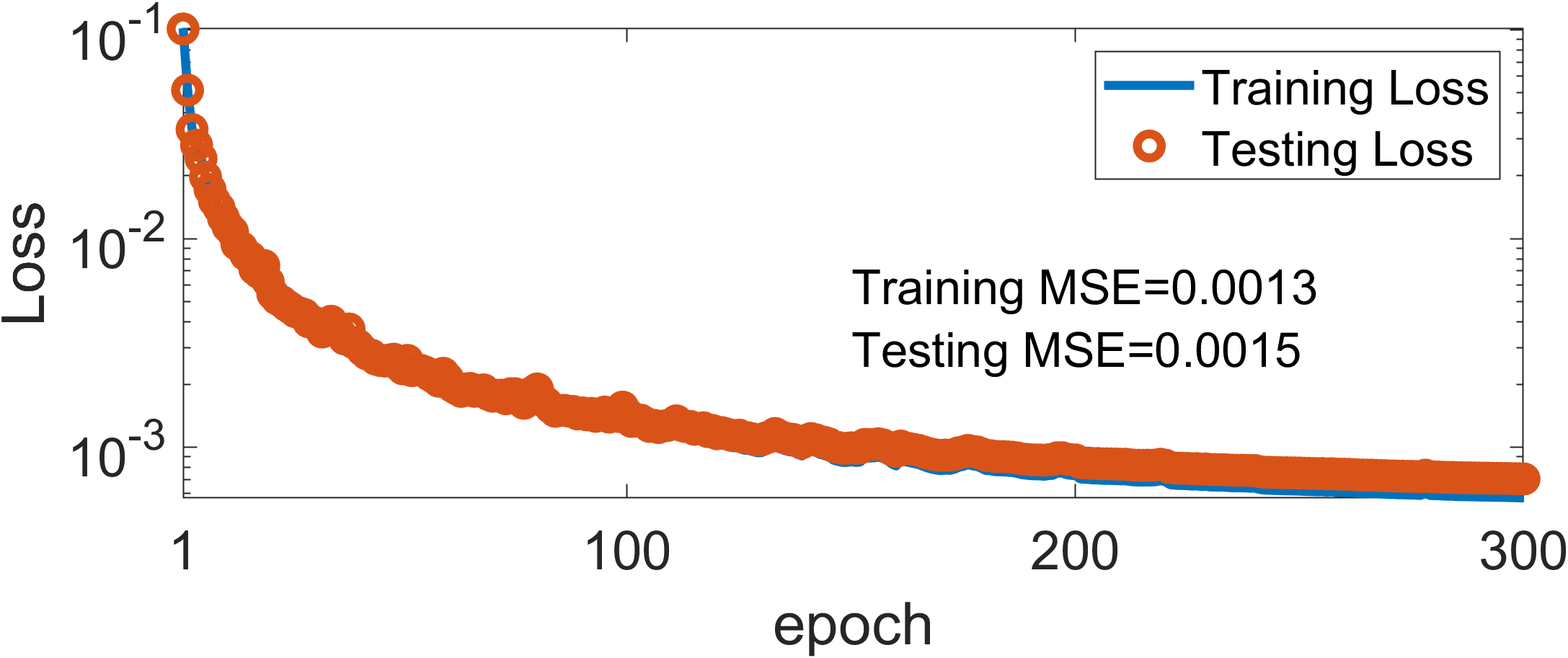}
	\caption{Convergence curve of GraphSolver for solving CFIEs of basic 3D targets.}
	\label{basicloss}
\end{figure}
\begin{figure}
	\centering
	\includegraphics[width=1\linewidth]{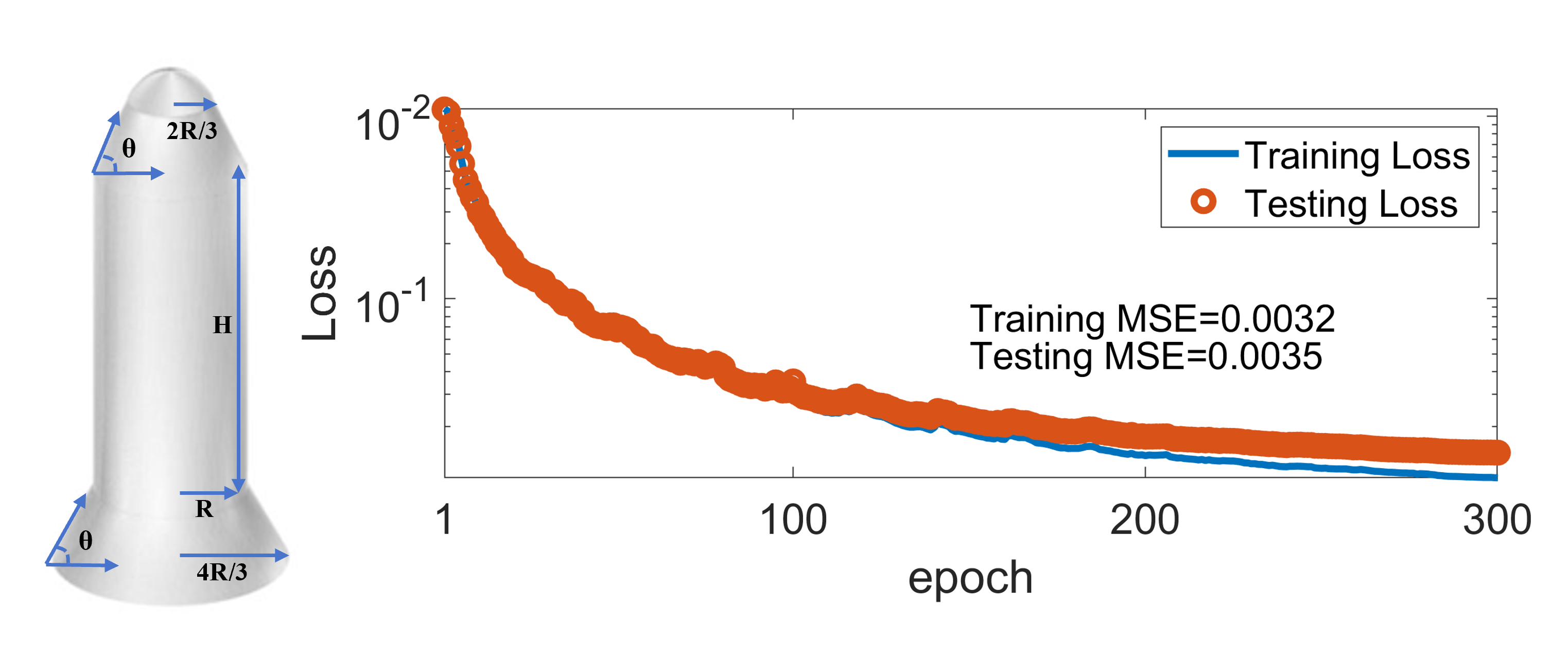}
	\caption{Schematic of missile-shaped targets and the convergence curve of GraphSolver for solving CFIEs of these targets.}
	\label{missleloss}
\end{figure}
\begin{figure}
	\centering
	\includegraphics[width=1\linewidth]{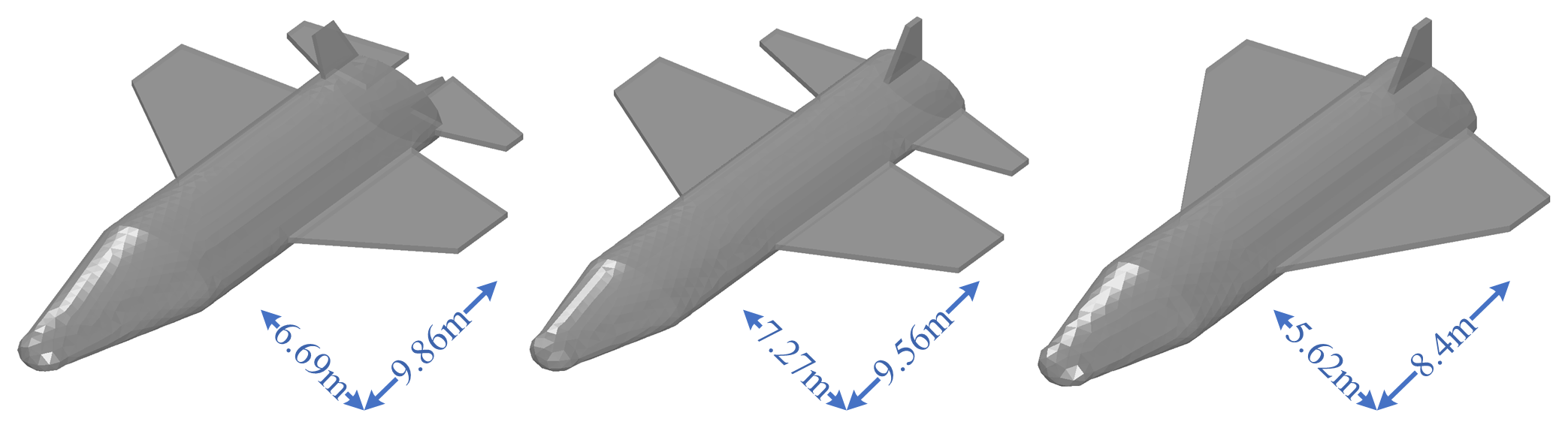}
	\caption{Schematic of three airplane-shaped targets with their respective lengths and widths indicated.}
	\label{airplane}
\end{figure}
\begin{figure}
	\centering
	\includegraphics[width=0.85\linewidth]{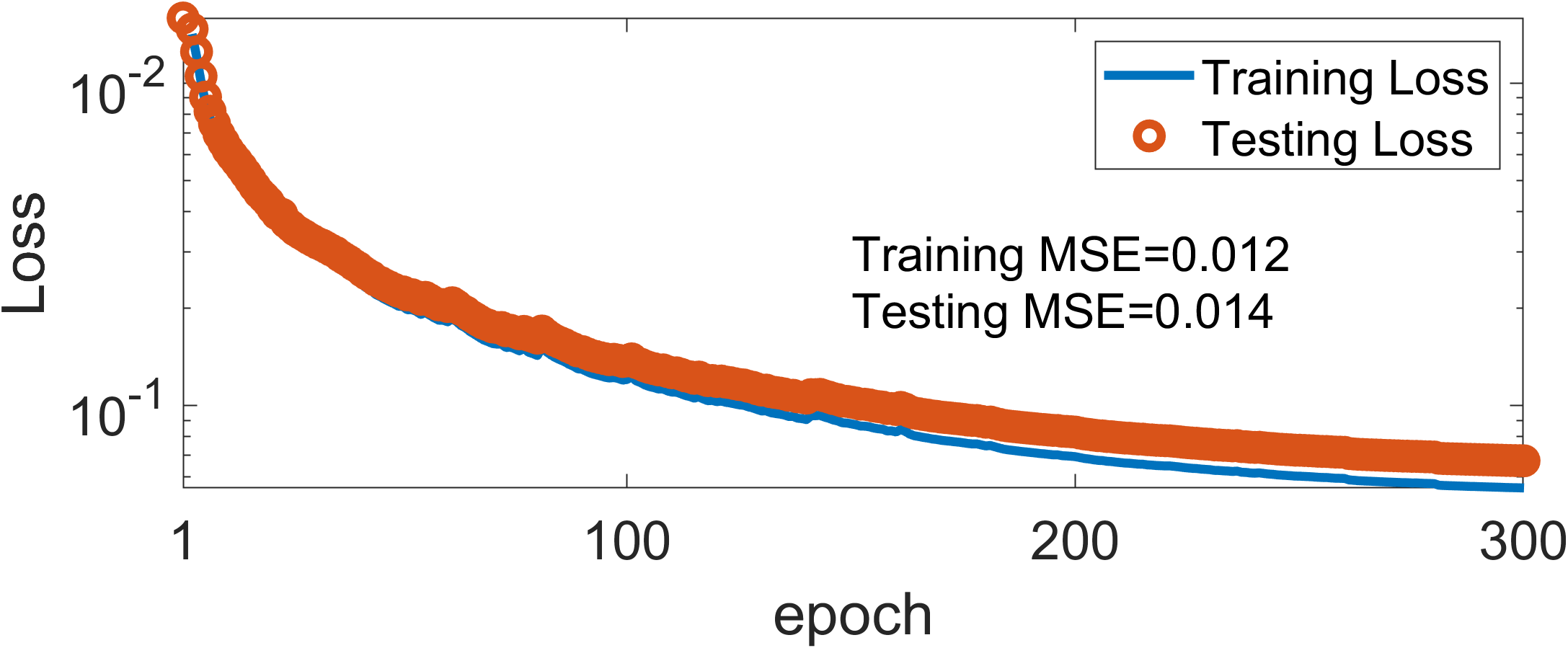}
	\caption{Convergence curve of GraphSolver for solving CFIEs of airplane-shaped targets.}
	\label{planeloss}
\end{figure}
\begin{figure*}
	\centering
	\includegraphics[width=0.95\linewidth]{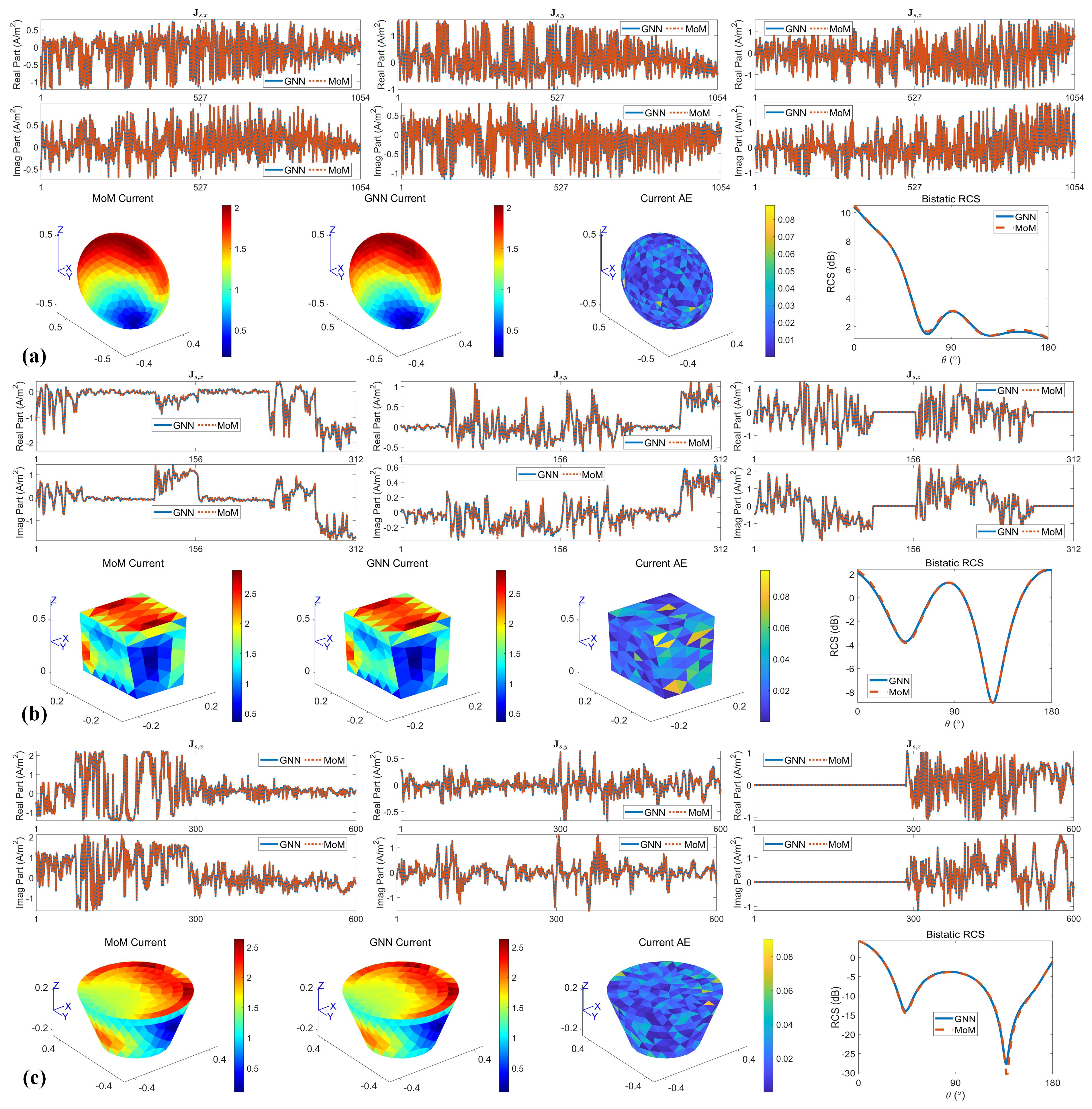}
	\caption{Comparison between the surface currents computed by MoM and GraphSolver. (a), (b) and (c) are the results of spheroid, conical frustum and hexahedron. In each sub-figure, the first row (from left to right) are the real parts of the $x$, $y$ and $y$ components of the surface current density, the second row (from left to right)  are the imaginary parts of the $x$, $y$ and $y$ components of the surface current density, the third row (from left to right) are 3D surface currents computed by MoM and GraphSolver, their AE distribution, bistatic RCS curves on the $\phi=0^{\circ}$ plane. }
	\label{basicresult}
\end{figure*}
\begin{table}
	\renewcommand{\arraystretch}{1.3}
	\caption{Control Parameters of Basic 3D targets and misslehead-shaped targets}
	\label{basicpara}
	\centering
	\begin{threeparttable}
		\begin{tabular}{ccc}
			\toprule
			spheriod  & conical frustum   &  hexahedron  \\
			\midrule
			Rx/Ry/Rz=$[0.2, 0.7]$ & Rt/Rz=$[0.2, 0.6]$ & Dt/Wt=$[0.2, 0.5]$  \\
			& H=$[0.2, 0.7]$  & Db/Wb/H=$[0.2, 0.6]$  \\
			\midrule
			& misslehead-shaped target &   \\
			\midrule
			$H=[1.1, 1.5]$ & $R=[0.2, 0.4]$& $\theta = [45^{\circ}, 70^{\circ}]$ \\
			\bottomrule
		\end{tabular}
	\end{threeparttable}
	\begin{tablenotes}
		\item[] $\theta$  is measured in angular units with increments of $5^{\circ}$ and all other parameters are expressed in meters with increments of $0.1$m.
	\end{tablenotes}
\end{table}
\begin{figure*}
	\centering
	\includegraphics[width=0.95\linewidth]{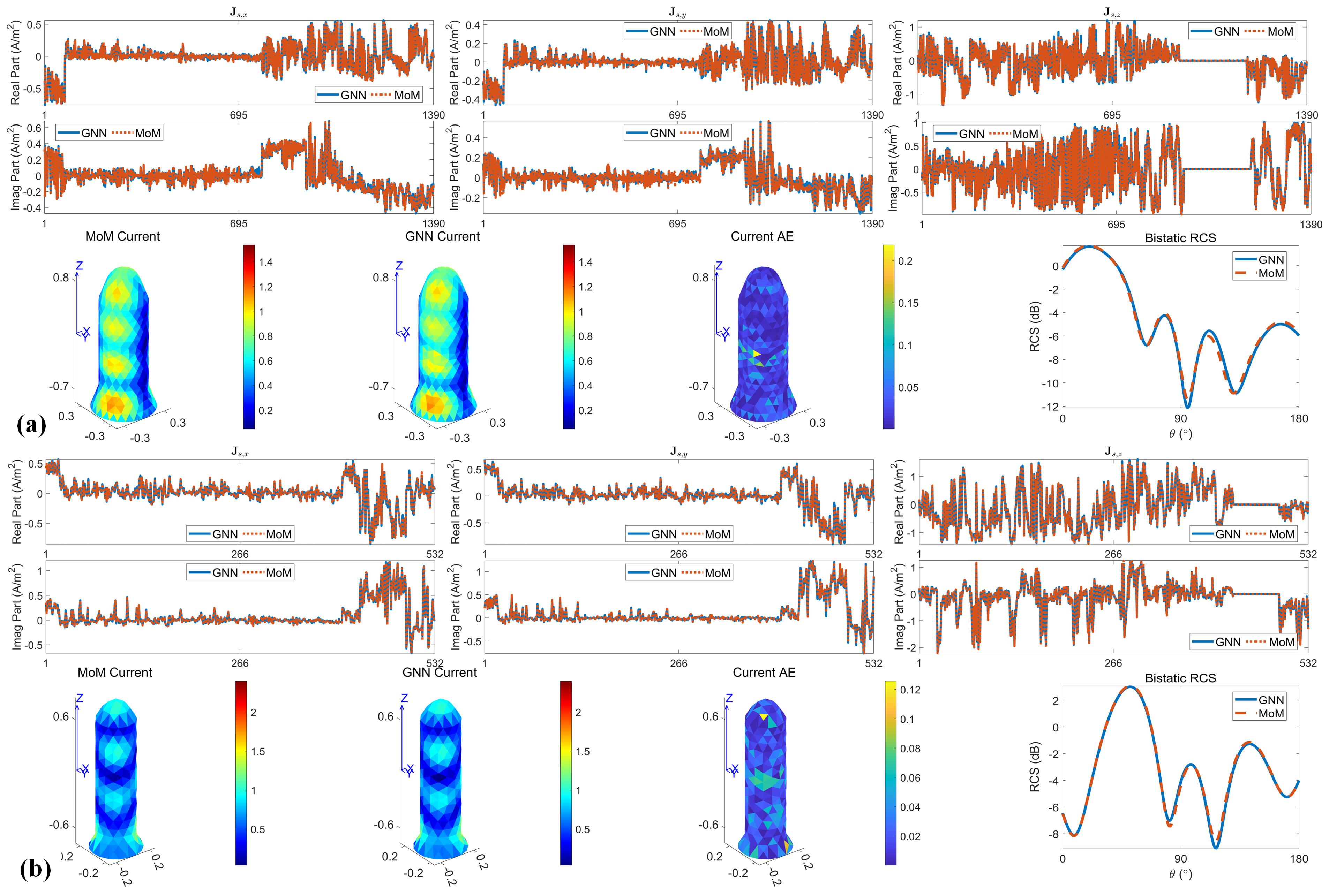}
	\caption{Comparison between the surface currents computed by MoM and GraphSolver. (a) and (b) are two different misslehead-shaped targets. In each sub-figure, the first row (from left to right) are the real parts of the $x$, $y$ and $y$ components of the surface current density, the second row (from left to right)  are the imaginary parts of the $x$, $y$ and $y$ components of the surface current density, the third row (from left to right) are 3D surface currents computed by MoM and GraphSolver, their AE distribution, bistatic RCS curves on the $\phi=0^{\circ}$ plane. }
	\label{missleresult}
\end{figure*}
\par 
GraphSolver adopts the supervised learning scheme by applying MoM to generate training data. The mean squared error (MSE) is adopted as the objective function to guide the training process:
\begin{equation}
	\text{MSE} = \frac{ ||\mathbf{J}_{s,m}-\mathbf{J}_{s,g}||_F^2 }{N_{\mathbf{J}_{s,m}}}
\end{equation}
where $\mathbf{J}_{s,m}$, $\mathbf{J}_{s,g}$ denote the surface current densities computed by MoM and GraphSolver, $N_{\mathbf{J}_{s,m}}$ is the total number of elements in $\mathbf{J}_{s,m}$, $||\cdot||_F$ is the Frobenius norm.
\begin{figure*}
	\centering
	\includegraphics[width=0.95\linewidth]{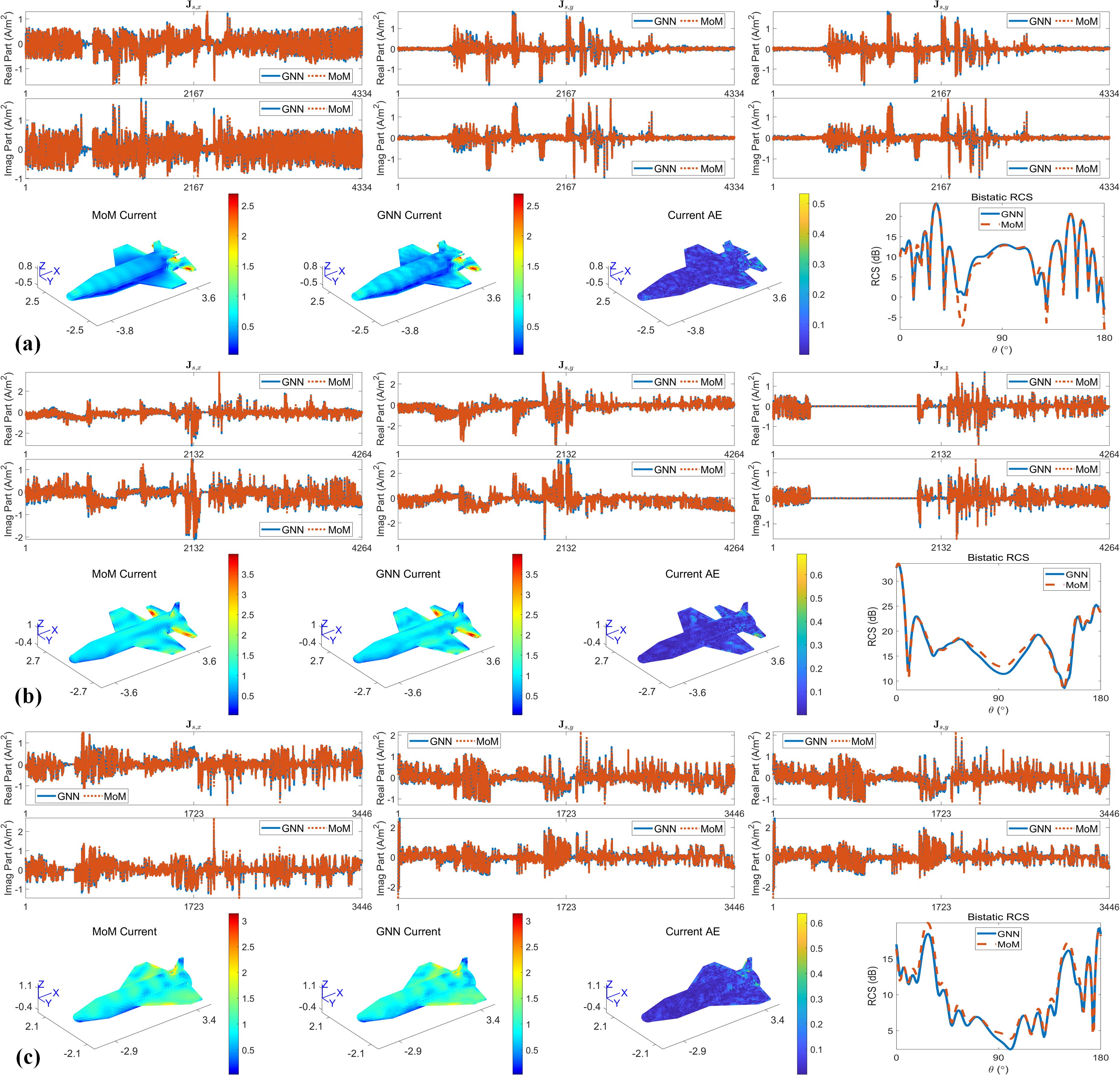}
	\caption{Comparison between the surface currents computed by MoM and GraphSolver. (a), (b) and (c) are the results of three different airplane-shaped targets. In each sub-figure, the first row (from left to right) are the real parts of the $x$, $y$ and $y$ components of the surface current density, the second row (from left to right)  are the imaginary parts of the $x$, $y$ and $y$ components of the surface current density, the third row (from left to right) are 3D surface currents computed by MoM and GraphSolver, their AE distribution, bistatic RCS curves on the $\phi=0^{\circ}$ plane. }
	\label{planeresult}
\end{figure*}
\section{Numerical Results}
In this section, the efficacy of GraphSolver is demonstrated through the solution of CFIEs for 3D conducting bodies exhibiting varying degrees of geometric complexity. 
The considered geometries, arranged in ascending order of complexity, include basic 3D targets, missile-shaped targets, and airplane-shaped targets. 
GraphSolver is independently trained to predict the current densities induced on these targets when illuminated by vertically polarized plane waves under varying incident angles. 
The implementation of GraphSolver utilizes PyTorch and is executed on an NVIDIA V100 GPU. The model parameters are optimized using the Adam optimizer.
\subsection{Basic 3D Targets}
The basic 3D targets considered in this study include spheroids, conical frustums, and hexahedrons, as illustrated in \fig{basictargets}.
The body centers of the spheroids and the base centers of the conical frustums and hexahedrons are aligned with the origin of the coordinate system. 
Their geometric configurations vary, with the corresponding control parameters detailed in \tab{basicpara}.
The illuminating plane wave has an amplitude of $1\text{V/m}$ and a frequency of $300\text{MHz}$. The incident direction, specified by $[\theta, \phi]$, varies within $[10^\circ, 90^\circ]$ for $\theta$ and $[90^\circ, 180^\circ]$ for $\phi$, with increments of $10^\circ$. 
MoM is employed to generate a dataset comprising 32400 data samples, with 80\% allocated for training and 20\% for testing.
To maintain the numerical stability, the average size of the RWG basis functions is set to $\lambda/10$, where $\lambda$ represents the wavelength of the incident plane wave.
\par 
\fig{basicloss} depicts the convergence curve of GraphSolver in solving CFIEs for basic 3D targets. The learning rate of the Adam optimizer is initialized at $0.001$ and reduced by a factor of $0.8$ every 20 epochs. GraphSolver undergoes training for a total of 300 epochs. Throughout the training process, the training and testing MSE values exhibit strong consistency. The converged MSE values are nearly identical, with minimal differences, indicating negligible overfitting, as highlighted in \fig{basicloss}.
\par
\fig{basicresult} presents a comparison between the surface currents of spheroid, conical frustum, and hexahedron targets, as computed by MoM and GraphSolver. 
Each target features distinct distributions and varying quantities of triangular meshes, with the number of unknowns being $1054$, $312$ and $600$ in \fig{basicresult} (a), (b) and (c), respectively. 
The outputs of GraphSolver include the real and imaginary components of the surface current density in the x, y and z directions for specific target.
The surface current densities of the various basic 3D targets exhibit distinct distributions and dynamic trends, as illustrated in \fig{basicresult}. 
Some curves display pronounced oscillations over a wide range, others oscillate within a narrow range, while certain cases result in flat, zero-valued lines. 
Despite these variations, GraphSolver consistently delivers stable and accurate predictions of surface current densities.
The corresponding 3D surface currents exhibit high precision, with absolute errors (AE) remaining at a low level.
To further validate performance, bistatic radar cross-section (RCS) curves are computed on the $\phi = 0^{\circ}$ plane for detailed comparisons.
The RCS curves derived from the surface currents computed by MoM and GraphSolver align closely for all three targets.
This demonstrates that GraphSolver effectively handles triangular meshes and underscores its strong potential for modeling 3D conducting bodies.
\subsection{Missilehead-shaped Targets}
In this section, GraphSolver is further trained to solve CFIEs for 3D conducting bodies with increased geometrical complexity.
\fig{missleloss} illustrates the schematic representation of the missilehead-shaped targets, with their geometrical parameters detailed in \tab{basicpara}.
The amplitude and frequency of the incident plane waves are set to $0.5$ and $300$MHz.
The incident direction $[\theta, \phi]$ varies within $[10^\circ, 90^\circ]$ and $[90^\circ, 180^\circ]$ with increments of $10^\circ$. 
MoM is employed to generate a total of $6750$ data samples of which 80\% and 20\% are for training and testing.
The average size of triangular meshes is also fixed as $\lambda/10$.
Adam optimizer is adopted to train GraphSolver, with the learning rate initialized at 0.001 and reduced by a factor of 0.8 every 20 epochs.
The convergence curve of GraphSolver for solving CFIEs of missilehead-shaped targets is also shown in \fig{missleloss}, where negligible overfitting is observed.
\par
\fig{missleresult} presents a comparison of the surface currents for two distinct missilehead-shaped targets, as computed by MoM and GraphSolver. 
The number of unknowns for the two targets is $1390$ and $532$ respectively.
The corresponding 3D surface currents and bistatic radar cross-section (RCS) curves on the $\phi=0^{\circ}$ plane demonstrate high computational accuracy.
The missilehead-shaped targets can be viewed as combinations of multiple basic 3D shapes. The resulting surface currents exhibit distinct and more complex distributions compared to those of basic 3D targets, as shown in \fig{missleresult}.
Despite this increased complexity, GraphSolver consistently delivers stable and accurate predictions of the surface currents, further validating its robust learning capabilities.
\begin{table}
	\renewcommand{\arraystretch}{1.3}
	\caption{Comparisons between PhiGRL and GraphSolver}
	\label{performance}
	\centering
	\begin{threeparttable}
		\begin{tabular}{ccc}
			\toprule
			\multicolumn{3}{c}{basic 3D targets} \\
			\midrule
			 \quad  & \quad \quad PhiGRL \quad \quad & \quad \quad GraphSolver \quad   \\ \hdashline[0.5pt/2pt]
			Training sample	& 25630  &  25630 \\
			Training epoch /time& 300 / $\approx$ 150h& 300 / $\approx$ 55h  \\
			Training/Testing MSE &$0.00064$/$0.00071$ &$0.0013$/$0.0015$\\
			\midrule
			\multicolumn{3}{c}{misslehead-shaped targets} \\
			\midrule
			& PhiGRL & GraphSolver \\ \hdashline[0.5pt/2pt]
			Training sample	&  5400 & 5400   \\
			Training epoch/time& 100 / $\approx$ 23h & 300 / $\approx$ 10h \\
			Training/Testing MSE & $0.0010$/$0.0011$ &$0.0032$/$0.0035$\\
			\midrule
			\multicolumn{3}{c}{airplane-shaped targets} \\
			\midrule
			& PhiGRL & GraphSolver \\ \hdashline[0.5pt/2pt]

			Training sample	& 5230  & 5230  \\
			Training epoch/time& 300 / $\approx$ 175h &  300 / $\approx$ 35h  \\
			Training/Testing MSE &$0.0050$/$0.0066$ &$0.012$ / $0.014$\\
			\bottomrule
		\end{tabular}
	\end{threeparttable}
\end{table}
\subsection{Airplane-shaped Targets}
In this section, GraphSolver is further trained to predict the surface currents of airplane-shaped targets using transfer learning. Transfer learning aims to transfer learned knowledge or physical laws from one domain to a similar domain, thereby simplifying the training process and reducing the need for large datasets\cite{zhuang2020comprehensive}.
The model parameter set of GraphSolver trained for 3D basic targets is taken as the starting point for training.
Airplane-shaped targets exhibit the most complex geometries in this study, featuring sharp corners, wing-body joints, and other challenging structures, which complicate accurate modeling, as shown in \fig{airplane}.
The fuselage length and wingspan of the airplane-shaped targets are also indicated. 
The frequency and amplitude of the incident plane waves are set to $150$MHz and $0.5$, respectively.
The direction of the incident plane waves, denoted by $(\theta, \phi)$, varies within the ranges $[0^{\circ}, 180^{\circ}]$ and $[3^{\circ}, 180^{\circ}]$ with increments of $5^{\circ}$ and $3^{\circ}$.
A total of $6660$ data samples are generated with MoM with the average size of RWG basises fixed as $\lambda/10$.
The training and testing data are split in an $80\%$-$20\%$ ratio.
The Adam optimizer is employed to optimize GraphSolver, with the learning rate initialized at $0.001$ and reduced by a factor of $0.8$ every 20 epochs. To account for the greater fluctuations in the surface current densities, GraphSolver incorporates batch normalization before the PReLU nonlinearities in all layers of GCN, as well as in all layers of FCNs except the last one.
\par 
\fig{planeloss} shows the convergence curve of GraphSolver for solving CFIEs of airplane-shaped targets. 
The number of training data samples is relatively insufficient given the geometric complexity of the targets, leading to slight overfitting and suboptimal numerical accuracy. 
Increasing the number of training samples could enhance GraphSolver’s performance. 
\fig{planeresult} compares the surface currents computed by MoM and GraphSolver. 
The x, y, and z components of the surface current densities for the airplane-shaped targets are more complex than those for basic 3D and missile-shaped targets.
Nonetheless, GraphSolver can still be trained to yield reliable predictions with acceptable numerical accuracy. The corresponding 3D surface currents are also computed and compared, showing a low level of absolute error. 
The bistatic RCS curves also demonstrate good agreement.
\subsection{Comparisons of Computing Performance}
In this section, we compare the performance of GraphSolver with that of PhiGRL, as reported in our previous work.
It is important to note that the version of PhiGRL considered here was trained using a supervised learning scheme.
PhiGRL and GraphSolver are built on different underlying principles, with the former belonging to the physics-inspired learning paradigm and the latter to the data-driven learning approach.
PhiGRL trains GNNs to learn update rules by numerically calculating the residuals of the governing equations, while GraphSolver consists of seven FCNs and one GCN, with no numerical computation involved.
\par 
\tab{performance} summarizes the training time and error levels of both GraphSolver and PhiGRL for different targets. In \cite{shan2023physics}, PhiGRL is trained to compute the coefficients of RWG basis functions, whereas GraphSolver predicts the surface current densities. To ensure a fair comparison, we also evaluate the average mean squared error (MSE) of the surface current densities derived from the coefficients computed by PhiGRL.
Both PhiGRL and GraphSolver were implemented on the same computing platform: a single Nvidia V100 GPU. While the incorporation of numerical computation in PhiGRL improves training stability and facilitates the learning task of the neural networks, it also increases computational overhead, requiring more training time. In contrast, GraphSolver achieves efficient training while maintaining comparable computational accuracy, making it more suitable for scenarios with limited computational resources.
\section{Conclusions}
In this paper, we present a fully data-driven fast solver to solve CFIEs of 3D conducting bodies by sequentially assembling one upsampling FCN, one GCN and six downsampling FCNs.
The 3D conducting bodies, discretized by RWG basis functions, are transformed into graphs in a concise and informative manner by treating each triangular patch as a node, thereby enabling the flow of current between nodes.
The efficacy of the proposed solver is verified by solving CFIEs for basic 3D targets, missile-shaped targets, and airplane-shaped targets. 
Numerical results show that the solver is able to handle different target shapes with varying degrees of geometric complexity, offering accurate predictions of surface current distributions.

Compared to physics-inspired learning methods, the data-driven solver facilitates efficient training and achieves acceptable computational accuracy, making it a strong candidate for scenarios with limited computational resources.
Despite some limitations related to training sample size and data quality, the proposed data-driven solver holds great promise for real-time applications where large-scale simulations are often computationally prohibitive.

\bibliographystyle{IEEEtran}
\bibliography{IEEEabrv,ref}

\end{document}